\pdfoutput=1
\documentclass[11pt]{article}

\usepackage{acl}

\usepackage{times}
\usepackage{latexsym}

\usepackage[T1]{fontenc}
\usepackage[utf8]{inputenc}

\usepackage{microtype}

\title{Unobserved Local Structures Make Compositional Generalization Hard}

\author{\makecell{Ben Bogin$^{1}$ ~~~~~~~ Shivanshu Gupta$^{2}$  ~~~~~ Jonathan Berant$^{1}$ } \\ 
$^{1}$Tel-Aviv University \hspace{4mm} $^{2}$University of California Irvine  \hspace{4mm}   \\
\texttt{\makecell{\{ben.bogin,joberant\}@cs.tau.ac.il, shivag5@uci.edu\\
}}}

\usepackage{graphicx}
\usepackage{amssymb}
\usepackage{amsmath}
\usepackage{booktabs}
\usepackage{makecell}
\usepackage{enumitem,kantlipsum}
\usepackage{hyperref}
\usepackage{algorithm,algorithmic}
\usepackage{arydshln}
\usepackage{caption}

\newif\ifcomments
\commentstrue
\ifcomments
    \providecommand{\bb}[1]{{\protect\color{olive}{[BB: #1]}}}
    \providecommand{\jb}[1]{{\protect\color{red}{[JB: #1]}}}
    \providecommand{\sg}[1]{{\protect\color{cyan}{[SG: #1]}}}
\else
    \providecommand{\bb}[1]{}
    \providecommand{\jb}[1]{}
    \providecommand{\sg}[1]{}
\fi

\newcommand{\tightparagraph}[1]{\smallbreak\noindent\textbf{#1}}
\definecolor{applegreen}{rgb}{0.01, 0.65, 0.01}
\definecolor{cardinal}{rgb}{0.77, 0.12, 0.23}

\begin{document}
\maketitle

\begin{abstract}

While recent work has shown that sequence-to-sequence models struggle to generalize to new compositions (termed \emph{compositional generalization}), little is known on what makes compositional generalization hard on a particular test instance. In this work, we investigate the factors that make generalization to certain test instances challenging. We first substantiate that some examples are more difficult than others by showing that different models consistently fail or succeed on the same test instances. Then, we propose a criterion for the difficulty of an example: a test instance is hard if it contains a \emph{local structure} that was not observed at training time. We formulate a simple decision rule based on this criterion and empirically show it predicts instance-level generalization well across 5 different semantic parsing datasets, substantially better than alternative decision rules. Last, we show local structures can be leveraged for creating difficult adversarial compositional splits and also to improve compositional generalization under limited training budgets by strategically selecting examples for the training set.

\end{abstract}
\section{Introduction}

Recent analyses of pre-trained sequence-to-sequence (seq2seq) models have revealed that they do not perform well in a \emph{compositional generalization} setup, i.e., when tested on structures that were not observed at training time \cite{lake2018scan,keysers2020measuring}.
However, while performance drops on average, models do not \emph{categorically} fail in compositional setups, and in fact are often able to successfully emit new unseen structures. This raises a natural question:  \textbf{What are the conditions under which compositional generalization occurs in seq2seq models?}

\begin{figure}
  \centering
  \includegraphics[width=0.8\linewidth]{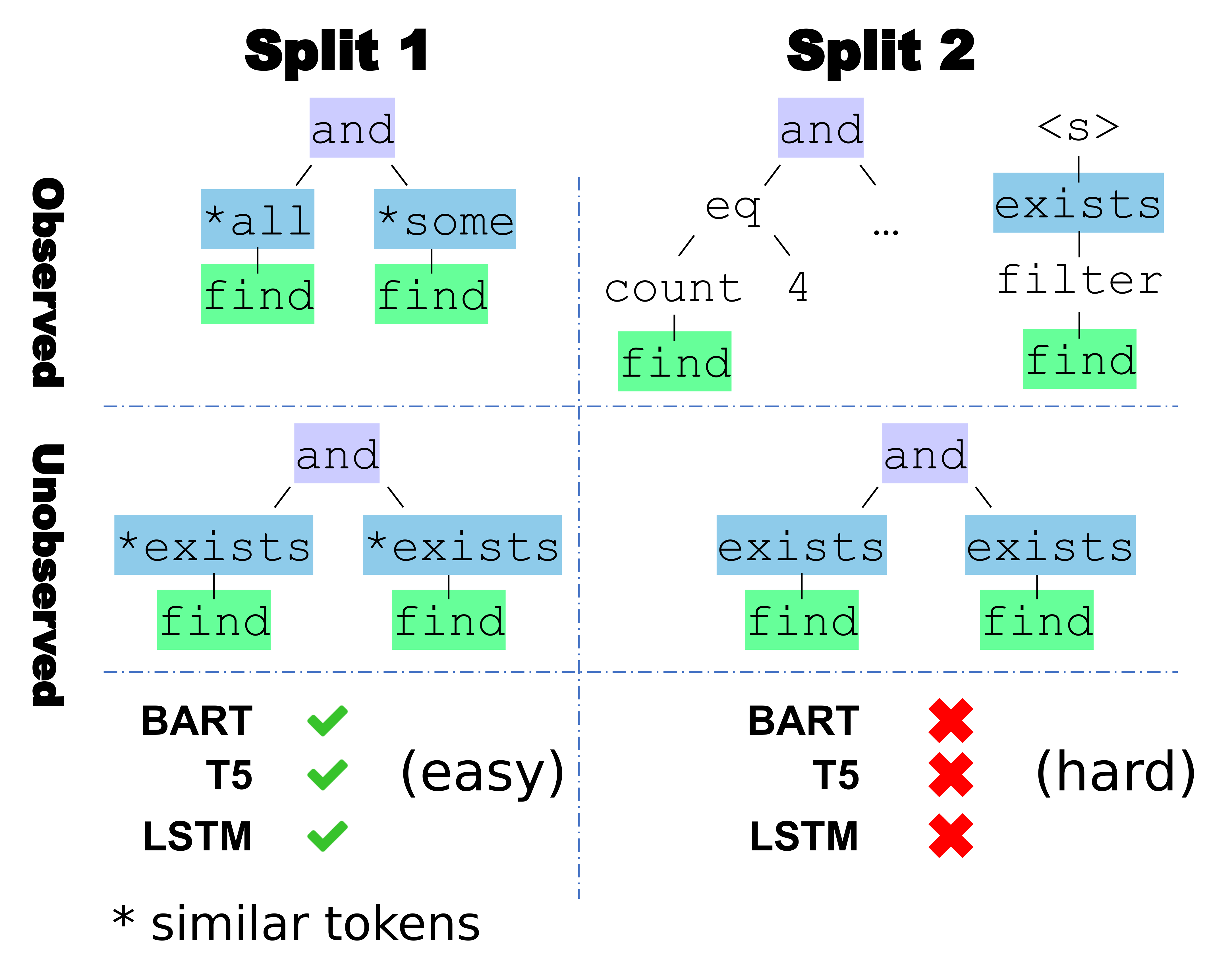}
  \caption{Unobserved local structures are harder for models to generalize to whenever there are no \emph{similar} structures that were observed during training.
  }
  \label{fig:intro}
\end{figure}

\begin{table*}[!t]
\centering
\scriptsize
\begin{tabular}{p{1.55cm}p{13.6cm}}
\toprule
{\bf Dataset} & {\bf Example} \\
 \midrule
  \makecell[l]{\textbf{COVR}\\(synthetic)} & \makecell[l]{\textit{What is the number of gray square cat that is looking at mouse?} \\ \texttt{count (with\_relation (filter (gray, filter (square, find(cat))), looking\_at, find (mouse)))}}\\
  \midrule
  \makecell[l]{\textbf{Overnight}\\(synthetic\textsuperscript{$\dagger$},\\
  natural\textsuperscript{$\circ$})
  } & \makecell[l]{\textsuperscript{$\dagger$}\textit{number of played games of player kobe bryant whose number of assists is 3}\\ \textsuperscript{$\circ$}\textit{how many games has kobe bryant made more than 3 assists} \\ \texttt{(listValue (getProperty (filter (getProperty kobe\_bryant (reverse } \\ \texttt{player)) num\_assists = 3) num\_games\_played))}}\\
  \midrule
  \makecell[l]{\textbf{Schema2QA}\\(synthetic)} & \makecell[l]{\textit{are there people that works for crossbars ?} \\ \texttt{( Person ) filter worksFor contains "crossbars"}}\\
  \midrule
  \makecell[l]{\textbf{ATIS}\\(natural)} & \makecell[l]{\textit{round trip flights between new york and miami} \\ \texttt{(lambda \$0 e (and (flight \$0) (round\_trip \$0) (from \$0 new\_york: ci) (to \$0 miami: ci)))}}\\
 \bottomrule
\end{tabular}
\caption{An example utterance-program pair for each of the datasets used in this work. 
}
\label{tab:datasets}
\end{table*}

Measuring compositional generalization at the 
\emph{dataset level}
obscures the fact that for a particular \emph{test instance}, performance depends on the instance sub-structures and the examples  observed during training. Consequently, it might be possible to predict how difficult a test instance is given the test instance and the training set. Indeed, papers that provide multiple compositional splits \cite{kim-linzen-2020-cogs,bogin-etal-2021-covr} have demonstrated high variance in accuracy across splits.
%, where some are easy to solve and some are much harder.

In this paper, we investigate the question of what makes compositional generalization hard in the context of semantic parsing, the task of mapping natural language utterances to executable programs. First, we create a large set of \emph{compositional} train/test data splits over multiple datasets, in which there is \emph{no} overlap between the programs of the test set and the training set. We then fine-tune and evaluate multiple seq2seq models on these splits. Our first finding is that different models tend to agree on which test examples are difficult and which are not. This indicates example difficulty can be mostly explained by the example itself and the training data, independent of the model. This calls for a better characterization of what makes a test instance hard.

To this end, we analyze the factors that make compositional generalization hard \emph{at the instance level}.
We formulate a simple decision rule that predicts the difficulty of test instances across multiple splits and datasets. Our main observation is that a test instance is hard if it contains a \emph{local structure} that was not observed at training time. An unobserved local structure is defined as a small connected sub-graph that occurs in the program of the test instance, but does not occur in the training set. Moreover, unobserved structures render an instance difficult only if there are no observed structures that are \emph{similar} to the unobserved one, where similarity is defined through a simple distributional similarity metric. 
Fig.~\ref{fig:intro} presents two splits that contain a tree with the same local structure in the test set.
Split 1 is easy because the training set contains similar local structures: \texttt{exists} is similar to \texttt{all} and \texttt{some}. Conversely, in Split 2 emitting the unobserved structure will be hard, as there are no similar observed structures in the training set.

We empirically evaluate our decision rule on five different datasets with diverse semantic formalisms, and show it predicts instance-level generalization well for both synthetic and natural language inputs, with an area under curve (AUC) score ranging from 78.4 to 93.3 (across datasets), substantially outperforming alternative rules.
%that do not consider the program structure.
Moreover, we compare our approach to MCD \cite{keysers2020measuring}, a metric that has been used to characterize difficulty at the dataset level (and \emph{not} the instance level). We show our rule can be generalized to the dataset level and outperforms MCD in predicting the difficulty of various compositional splits.
Last, we find that our decision rule applies not just to Transformers, but also to LSTM decoders.

With these insights, we use our decision rule for two purposes. First, we develop a method for creating difficult compositional splits by picking a set of similar local structures, and holding out instances that include any of these structures. 
We show that seq2seq models get much lower accuracy on these splits compared to prior approaches. Second, we propose a data-efficient approach for selecting training examples in a way that improves compositional generalization. Given a large pool of examples, we select a training set that maximizes the number of observed local structures. We show this leads to better compositional generalization for a fixed budget of training examples. We release code and data at \url{https://github.com/benbogin/unobserved-local-structures}.

\section{Setup}
\label{sec:setup}
We focus on semantic parsing, where the task is to parse an utterance $x$ into an executable program $z$. In a compositional generalization setup, examples are split into a training and a test set, such that there is no overlap between programs in the two sets.

We use two methods to generate compositional splits. The first is the \textbf{template split}, proposed in \citet{finegan-dollak-etal-2018-improving}, where examples are randomly split based on an abstract version of their programs. Additionally, we propose a \textbf{grammar split}, which can be used whenever we have access to the context-free grammar rules that generate the dataset programs. 
A grammar split is advantageous because, unlike the template split, it creates meaningful splits where particular structures are held out.
For details on these split methods see App.~\ref{app:splits}.

\tightparagraph{Datasets} We use five different datasets, covering a wide variety of domains and semantic formalisms. Importantly, since we focus on compositional generalization, we choose datasets for which a random, i.i.d split, yields \emph{high} accuracy. Thus, errors in the compositional setup can be attributed to the compositional challenge, and not to  conflating issues such as lexical alignments or ambiguity.

We consider both synthetic datasets, generated with a synchronous context-free grammar (SCFG) that generates utterance-program pairs, and datasets with natural language utterances. The datasets we use are (see Table~\ref{tab:datasets} for examples):

\noindent 
\textbf{COVR}: A synthetic dataset based on \citet{bogin-etal-2021-covr} with a variable-free functional language.

\noindent 
\textbf{Overnight} \cite{wang-etal-2015-building}: Contains both synthetic and natural utterances covering multiple domains; uses the Lambda-DCS formal language \cite{liang-etal-2011-learning}. Examples are generated using a SCFG, providing synthetic utterances, which are then manually paraphrased to natural language.
%through crowdsourcing.

\noindent 
\textbf{Schema2QA} (\textbf{S2Q}, \citealt{xu2020schema2qa}): Uses the ThingTalk language \cite{campagna2019genie}. We only use the synthetic examples from the \textit{people} domain, provided by \newcite{oren-etal-2021-finding}.

\noindent 
\textbf{ATIS} \cite{Hemphill1990atis,dahl-etal-1994-expanding}: A dataset with natural language questions about aviation, and $\lambda$-calculus as its formal language.
\section{Model Agreement across Examples}
\label{sec:difficulty}

Our goal is to predict the difficulty of test instances. However, first we must confirm that (1) splits contain both easy and difficult instances and (2) different models agree on how difficult a specific instance is. To check this, we evaluate different models on multiple splits.

\smallbreak
\noindent
\textbf{Splits} We experiment with the five datasets listed in \S\ref{sec:setup}. For COVR, we use all of the 124 grammar splits that were generated. For Overnight, we generate 5 template splits for each domain, using the same splits for the synthetic and the natural language versions. For ATIS and S2Q, we generate 5 template splits.
%Since the training sets of COVR splits varied in size, we limited them up to size 2500 for a fair comparison across splits. 
%The size of the training sets across datasets ranges from 600-18,000, however, for a single dataset (and in Overnight, a single domain), the size of the training set is roughly similar across splits. 
%For Overnight and ATIS, we hold out 20\% of all program templates. For Schema2QA, we hold out 70\%, since otherwise model performance is too high.
We provide further details in App.~\ref{app-sub:split-sizes}.

\begin{figure}
  \centering
  \includegraphics[width=0.75\linewidth]{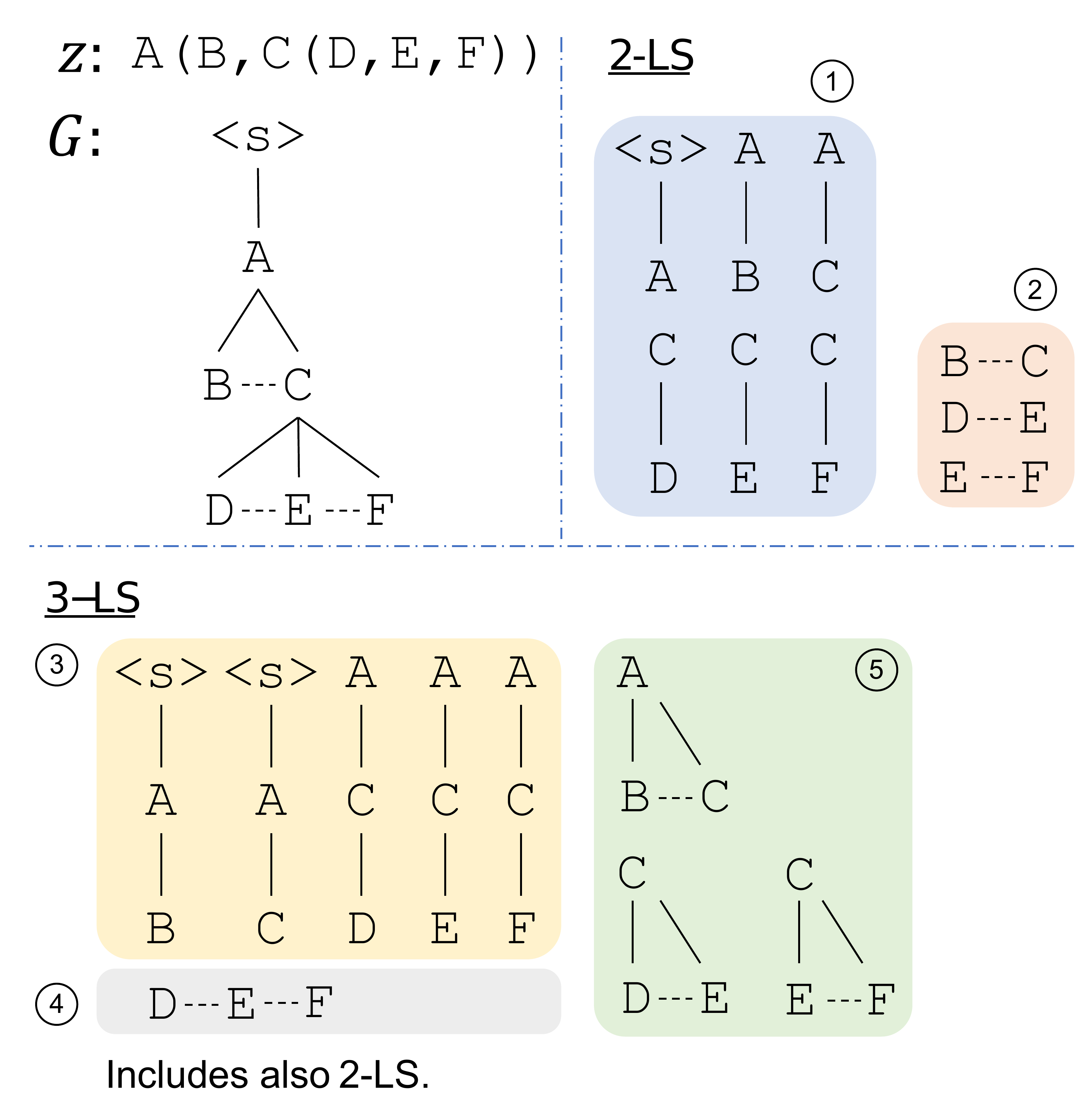}
  \caption{An example program $z$ and the structure of its program graph $G$ (top left), with solid edges for parent-child relations and dashed edges for consecutive siblings. In the other figure parts we enumerate all 2-LS and 3-LS structures over this graph.}
  \label{fig:local-structures}
\end{figure}

\tightparagraph{Models and experimental setting}
We experiment with four pre-trained seq2seq models: T5 Base/Large and BART-Base/Large \cite{Raffel2020t5,lewis-etal-2020-bart} and fine-tune them on each of the splits. See App.~\ref{app:training} for training details.

We evaluate performance with \emph{exact match (EM)}, that is, whether the predicted output is equal to the gold output. However, since EM is often too strict and results in false-negatives (i.e. a predicted target could be different than the gold target but still semantically equivalent), we use a few manually-defined relaxations (detailed in App.~\ref{app:datasets}). This is important for our analysis since we want to focus on meaningful failures rather than evaluation failures.

\tightparagraph{Results}
Table~\ref{tab:splits-accuracies} (top) shows average results across splits. We see that the accuracy of different models is roughly similar across datasets. Models tend to do well on most compositional splits, except in Overnight with natural language inputs.
To measure agreement, we compute \emph{agreement rate}, that is, the fraction of times all (or most) models either output a correct prediction or output an erroneous one. We observe that agreement rate is high (Table~\ref{tab:splits-accuracies}, middle): at least 3 models agree (3+/4) in 93.1\%-96.4\% of the cases, and all (4/4) models agree in 75.3\%-86.8\%. We compare
this to \emph{random agreement rate} (Table~\ref{tab:splits-accuracies}, bottom), where we compute agreement rate assuming that a model with accuracy $p$ outputs a correct prediction for a random subset of $p$ the examples and an incorrect prediction for the rest. Agreement rate is dramatically higher than random agreement rate.

\begin{table}[!t]
\centering
\scriptsize
\begin{tabular}{lllll}
\toprule
{\bf Model} & {\bf COVR} & {\bf Overnight} & {\bf S2Q} & {\bf ATIS} \\
&  & syn. / nat. & & \\
 \midrule
  T5-Base & 88.2 & 59.0 / 23.6 & 83.4 & 78.4  \\
  T5-Large & 87.8 & 62.3 / 27.4 & 88.0 & 77.6  \\
  BART-Base & 86.3 & 60.4 / 27.8 & 77.9 & 76.0 \\
  BART-Large & 85.4 & 60.3 / 28.2 & 82.0 & 78.9 \\
 \midrule
  Agree. (3+/4) & 94.5 & 96.2 / 93.5 & 93.1 & 96.4 \\
  Agree. (4/4) & 82.5 & 86.8 / 76.9 & 75.3 & 85.6 \\
 \midrule
 Rnd-agree. (4/4) & 57.1 & 15.8 / 29.3 & 47.0 & 36.7 \\
 \bottomrule
\end{tabular}
\caption{Average test EM across splits for all datasets, agreement rate for at least 3 models (3+/4) and all 4 models (4/4), and random agreement rate (4/4). For Overnight, we show results both on the synthetic and the natural language settings.}
\label{tab:splits-accuracies}
\end{table}

Importantly, the fact that models have high agreement rate suggests that instance difficulty depends mostly on the instance itself and the training set, and not on the model.

\section{What makes an instance hard?}

The results in \S\ref{sec:difficulty} provide a set of test instances of various difficulties from a variety of compositional splits.
We analyze these instances and propose a hypothesis for what makes generalization to a test instance difficult.

\subsection{Unobserved Local Structure}
\label{subsec:structures}

We conjecture that a central factor in determining whether a test instance is difficult, is whether its program contains any \textbf{unobserved local structures}. We represent a program as a graph, and a local structure as a connected sub-graph within it. We formally define these concepts next.

The output $z$ in our setting is a sequence of tokens which defines a program. Each token in $z$ represents a program \emph{symbol}, which is a function or a value, except for structure tokens (namely, parentheses and commas) that define parent-child relations between function symbols and their arguments. We parse $z$ into a tree $T=(\mathcal{V},\mathcal{E})$, such that each node $v \in \mathcal{V}$ is labeled by the symbol it represents in $z$,
and the set of edges $\mathcal{E}=\{(p,c)\}$ expresses \textbf{parent-child} relations between the nodes. We additionally add a root node \texttt{<s>} connected as a parent to the original root in $T$.

To capture also sibling relations, we define a graph based on the tree $T$ that contains an edge set $\mathcal{E}_{\text{sib}}$ of \textbf{sibling} edges: $G=(\mathcal{V},\mathcal{E} \cup \mathcal{E}_{\text{sib}})$. Specifically,
for each parent node $p$, the program $z$ induces an order over the children of $p$: $(c^p_1, ..., c^p_{N_p})$, where $N_p$ is the number of children. We then define $\mathcal{E}_{\text{sib}}=\bigcup_p\{c_i^p, c_{i+1}^p\}_{i=1}^{N_p}$, that is, all \emph{consecutive} siblings will be connected by edges. Fig.~\ref{fig:local-structures} (left) shows an example  program $z$ and its graph $G$.

We define local structures as connected sub-graphs in $G$, with $2 \leq n \leq 4$ nodes, that have a particular structure, presented next. We term a local structure with $n$ nodes as $n$-LS.
The set of 2-LSs refers to all pairs of parents and their children, and all pairs of consecutive siblings (Fig.~\ref{fig:local-structures}, structures 1 and 2). The set of 3-LSs includes all 2-LSs, and also structures with (1) two parent-child relations, (2) two siblings relations and (3) a parent with two siblings (structures 3, 4 and 5 in the figure, respectively). The structure 4-LS is a natural extension of 2-LS and 3-LS, defined in App.~\ref{app:4-ls}. 
Importantly, the structures we consider are local since they are \emph{connected}: We do not consider, for example, a grandparent-grandchild pair, or non-consecutive siblings.

\subsection{Similarity of Local Structures}
\label{subsec:similarity}

Our hypothesis is that if a model observes a test instance with an unobserved local structure, this instance will be difficult. We relax this hypothesis and propose that an example might be easy even if it contains an unobserved local structure $s_1$, if the training set contains a \emph{similar} structure $s_2$.

The similarity $\textrm{sim}(s_1, s_2)$ of two structures is defined as follows. Two structures can have positive similarity if (1) they are \emph{isomorphic}, that is, they have the same number of nodes and the same types of edges between the nodes, and (2) if they are identical up to flipping the label of a \emph{single} node. In any other case, similarity will be 0.0. If all nodes are identical, similarity is 1.0. When the structures differ by the label of a single node, we define their similarity using the \emph{symbol similarity} of the two symbols $m_1$ and $m_2$ that are different, which is computed using a distributional similarity metric $\widehat{\textrm{sim}}(m_1, m_2)$.

\begin{figure}
  \centering
  \includegraphics[width=1\linewidth]{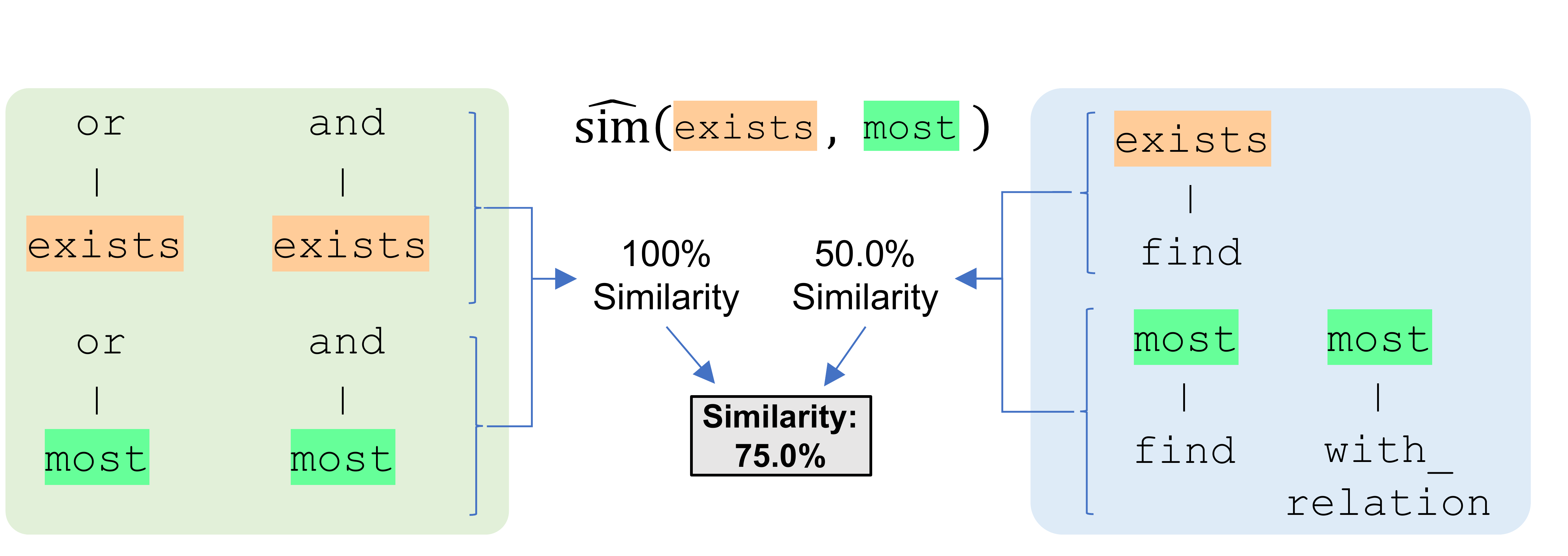}
  \caption{Example for the computation of similarity $\widehat{\textrm{sim}}(m_1, m_2)$ between two symbols $m_1,m_2$.}
  \label{fig:similarity}
\end{figure}

To compute symbol similarity we use the set of programs $\mathcal{P}_o$ in all training examples. We use this set to find the \emph{context} that co-occurs with each symbol -- that is, the set of symbols that have appeared as parents, children or siblings of a given symbol $m$. Specifically, we consider four types of contexts $c \in \mathcal{C}$, including children, parents, left siblings, and right siblings. We define $\textrm{ctx}_c(m)$ as the set of symbols that have appeared in the context $c$ of the symbol $m$. Finally, given two symbols $m_1$ and $m_2$, we average the Jaccard similarity across the set of \emph{relevant} context types $\tilde{\mathcal{C}} \subseteq \mathcal{C}$:
\begin{equation*}
    \widehat{\textrm{sim}}(m_1, m_2) = \frac{1}{|\tilde{\mathcal{C}}|} \sum_{c \in \tilde{\mathcal{C}}} {\textrm{jac}(\textrm{ctx}_c(m_1), \textrm{ctx}_c(m_2))},
\end{equation*}
where $\tilde{\mathcal{C}}$ contains a context type $c$ iff the set of context symbols $\textrm{ctx}_c(\cdot)$ is not empty for $m_1$ or $m_2$.

Consider the top example in Fig.~\ref{fig:similarity}. The token  \texttt{exists} appears with 2 different parents: \texttt{or} and \texttt{and}. The token \texttt{most} appears with exactly the same set of parents, thus their ``parent'' similarity is 100\%. Likewise, the ``children'' similarity of the two is 50\% since they share 1 out of 2 distinct children (\texttt{find}). Finally, the similarity between the tokens is the average of the two types of similarities, 75.0\% (for brevity, sibling contexts are not considered in the figure).

\subsection{Decision Rule}
We can now use the similarity between structures to predict the difficulty of an unobserved program $p_u$ given a set of observed programs $\mathcal{P}_o$. For exposition purposes, instead of predicting  \emph{difficulty}, we predict its complement: \emph{easiness}.

We compute the easiness of a program $p_u$ by comparing its local structures with the local structures in $\mathcal{P}_o$. Thus, we extract from $p_u$ the set $\mathcal{S}_u$ of $n$-LSs as defined in \S\ref{subsec:structures}, for a chosen $n$. Similarly, we extract the set $\mathcal{S}_o$ of all $n$-LSs in the set of training programs $\mathcal{P}_o$, and define the easiness:
%\jb{I think first show the formula and then explain it in a way that mirros the formula, if it's easyness let's stick to that}
\begin{equation*}
    \hat{e}(p_u) = \min_{s_u \in \mathcal{S}_u} \max_{s_o \in \mathcal{S}_o} \left( \textrm{sim} ( s_u, s_o ) \right),
\end{equation*}
that is, the easiness of $p_u$ is determined based on the \emph{least easy} (most difficult) unobserved structure in $\mathcal{S}_u$. The easiness of a particular structure $s_u$ is determined by the structure in  $\mathcal{S}_o$ that is \emph{most similar} to it.

\subsection{Alternative Decision Rules}
\label{subsec:alternative}

We discuss alternative decision rules, which will be evaluated as baselines in \S\ref{sec:experiments}.

\tightparagraph{N-grams over sequence} While we define structures over program \emph{graphs}, seq2seq models observe inputs and outputs as a flat sequence of symbols. Thus, it is possible that unobserved \emph{n-grams} in test instances, rather than local structures, explains difficulty. To test this, we define our decision rule to be identical, but replace local structures with consecutive n-grams in the program sequence, and consider only two context types (left/right co-occurrence) to compute symbol similarity.

\tightparagraph{Length} It is plausible that long sequences are more difficult to generalize to. To test this, given a set of training programs $\mathcal{P}_o$, we measure the number of symbols $m_l$ in the longest program in $\mathcal{P}_o$, and define the easiness of a program $p_u$ of length $m_u$ to be $\max \left( 1-\frac{m_u}{m_l}, 0 \right) $.

\tightparagraph{TMCD} 
The MCD and TMCD methods \cite{keysers2020measuring,shaw-etal-2021-compositional} have been used to create compositional splits, by maximizing \emph{compound divergence} across the training and test splits. A compound, which is analogous to our $n$-LS, is defined as any sub-graph of up to a certain size in the program tree, and divergence is computed over the distributions of compounds across the two sets (see papers for details).
We can use this method also to \emph{predict} difficulty instead of creating splits and compare it to our approach. While the two methods are not directly comparable, since we focus on \emph{instance-level generalization}, we extend our approach for computing the easiness of \emph{a split} and compare to TMCD in §\ref{sec:experiments}.

\section{Experiments}
\label{sec:experiments}
We now empirically test how well our decision rule predicts the easiness of test instances.

\tightparagraph{Setup}
We formalize our setup as a binary classification task where we predict the easiness $\hat{e}(p_u) \in [0,1]$ of a test instance (utterance-program pair) with program $p_u$, and compare it to the ``gold'' easiness $e(p_u) \in \{0,1\}$, as defined next. For each test instance, we have the EM accuracy of four models (\S\ref{sec:difficulty}). We thus define $e(p_u)$ to be the \emph{majority} EM on $p_u$ between the four models. We discard examples with no majority (3.6\%-6.9\% of the cases). In each dataset, we combine all test instances across all splits. We evaluate with Area Under the Curve (AUC), a standard metric for binary classification that does not require setting a threshold, where we compute the area under the curve of the true positive rate against the false positive rate.

\tightparagraph{Decision rules}
We compare variations of our $n$-LS decision rule with $n\in\{2,3,4\}$. Additionally, we evaluate the N-grams over sequence baseline (\textsc{2-Bigram}) and \textsc{Length}. The \textsc{Random} baseline samples a number between 0 to  1 uniformly.

We also conduct three ablations. The first, \textsc{2-LS-NoSib} ignores sibling relations.
%This affects not only the set of local structures considered, but also the context types used in the symbol similarity function.
Similarly, \textsc{2-LS-NoPC} ignores parent-child relations. Last, \textsc{2-LS-NoSim} tests a more strict decision rule that ignores structure similarity, i.e., $\hat{e}(p_u)=0$ for a test program $p_u$ that has \emph{any} unobserved 2-LS, even if the unobserved structures have similar observed structures in the training set.

\begin{table}[!t]
\centering
\scriptsize
\begin{tabular}{lllll}
\toprule
{\bf Decision Rule}       & {\bf COVR}    & {\bf Overnight}  & {\bf S2Q}    & {\bf ATIS} \\
&  & syn. / nat. & & \\
 \midrule
  \textsc{2-LS} & \textbf{93.3} & 84.6 / 67.9           & 78.9          & 75.8  \\
  \textsc{3-LS} & 93.1          & \textbf{91.0} / 74.8  & \textbf{81.6} & 78.7  \\
  \textsc{4-LS} & 92.1          & 88.0 / \textbf{78.4}  & 79.9          & \textbf{81.4} \\
 \midrule
  \textsc{2-LS-NoSib}   & 91.9  & 78.0 / 65.8           & 75.9          & 67.7 \\
  \textsc{2-LS-NoPC}    & 77.2  & 69.6 / 59.7           & 61.0          & 64.1 \\
  \textsc{2-LS-NoSim}   & 85.1  & 82.3 / 67.2           & 79.0          & 73.5 \\
  \textsc{2-Bigram}  & 88.5  & 58.2 / 52.4           & 69.7          & 69.9 \\
  \textsc{Length}       & 49.7  & 45.9 / 55.3           & 43.4          & 47.4 \\
  \textsc{Random}       & 50.5  & 49.5 / 51.6           & 48.2          & 49.2 \\
 \bottomrule
\end{tabular}
\caption{AUC scores of different decision rules for each dataset, computed across the test instances in all splits.}
\label{tab:decison-rule-results}
\end{table}

\begin{table*}[!t]
\centering
\scriptsize
\begin{tabular}{ll}
\toprule
Utterance & Either the number of dog is greater than the number of white animal or there is dog \\
\midrule
Gold & \texttt{{\color{applegreen}or(gt(count(find(dog)),count(filter(white,find(animal)))),exists(find(dog)))}} \\
\midrule
T5 & \texttt{{\color{applegreen}or(gt(count(find(dog)),count(filter(white,find(animal)))),}{\color{cardinal}there(find(dog)))}} \\
BART & \texttt{{\color{applegreen}or(gt(count(find(dog)),count(filter(white,find(animal)))),}{\color{cardinal}ists(find(dog)))}} \\
T5-L & \texttt{{\color{applegreen}or(gt(count(find(dog)),count(filter(white,find(animal)))),}{\color{cardinal}eq(find(dog)))}} \\
BART-L & \texttt{{\color{applegreen}or(gt(count(find(dog)),count(filter(white,find(animal)))),}{\color{cardinal}with\_relation(...}} \\
 \bottomrule
\end{tabular}
\caption{Example predictions of four models, showing a typical case where all models emit wrong tokens exactly when encountering an unobserved parent-child structure: \texttt{or}-\texttt{exists} (\texttt{L} in model name stands for large version).}
\label{tab:token-error}
\end{table*}
% split 100

\subsection{Results}
\label{subsec:results}
AUC scores for all decision rules are in Table~\ref{tab:decison-rule-results}, showing that our $n$-LS classifiers get high AUC scores, ranging from 78.4-93.3, outperforming the baselines and ablations.

Comparing performance across the order $n$ of LSs, there is some variance across datasets, and the best $n$ may depend on the dataset. Still, local structure explains generalization better than the baselines practically for all $n$'s.

The two graph-relation ablations (\textsc{2-LS-NoSib} and \textsc{2-LS-NoPC}) show that parent-child relations are more important than sibling relations, but both contribute to the final easiness score. The strict similarity ablation suggests that considering similarity between structures is important in COVR, but less in Overnight and ATIS, and not at all in S2Q. 
We hypothesize that similarity is important in COVR since the splits were created with a grammar, and not with templates. In grammar splits, often a single group of similar structures is split across the training and test sets (e.g., half of the quantifiers are in the training set, and half in the test set). In such cases, considering local structure similarity is important, since such test instances are easier according to our conjecture. %Such splits are unlikely to emerge in a template split.
The experiment we conduct in \S\ref{subsec:n-ls-splits}, where we specifically create such splits, supports this claim.

The low accuracy of \textsc{2-Bigram} indicates that unobserved structures in program space are better predictors compared to unobserved sequences. This is interesting since models train with symbol sequences. Last, we see that predicting difficulty by example length is as bad as a random predictor.

Performace on datasets with natural language (ATIS and Overnight-paraphrased) are lower than synthetic datasets. One reason is that natural language introduces additional challenges such as ambiguity, lexical alignment and evaluation errors -- mistakes that cannot be explained by our decision rule. 
We further discuss limitations in~\S\ref{sec:limitations}.

\tightparagraph{Token-level analysis} 
%We analyzed model errors at the instance level. Next, we analyze errors at the \emph{token level}.  Specifically, 
We now analyze the relation between the first incorrect token the models emit and the unobserved structures. Consider the example in Table~\ref{tab:token-error}: given a test example with an unobserved parent-child structure \texttt{or}-\texttt{exists}, all models emit a wrong token precisely where they should have output the child, \texttt{exists}. 

We measure the frequency of this by looking at model outputs where (1) the model is wrong and (2) unobserved \textsc{2-LS}s $\mathcal{S}_u$ were found in the gold program. For each such model output $\hat{z}$ and gold output $z$, we find the index $i$ of the first token where $\hat{z}^i \neq z^i$. We then count the fraction of cases where there is an unobserved 2-LS (a pair of symbols) $(m_1,m_2) \in \mathcal{S}_u$ such that both $m_1$ appears in the program prefix $\{z\}_{j=1}^{i-1}$ and $m_2 = z^i$. For COVR, this happens in 76.5\% of the cases. For Overnight, S2Q and ATIS, this happens in 28.4\%, 31.8\% and 45.1\% of the cases respectively, providing strong evidence that models struggle to emit local structures they have not observed during training.

\subsection{Comparison to TMCD}
\label{subsec:mcd}

\begin{table}[!t]
\centering
\scriptsize
\begin{tabular}{lll}
\toprule
{\bf Split}       & {\bf COVR}  & {\bf Overnight} \\
 \midrule
  \textsc{2-LS}   & \textbf{0.79} & 0.84  \\
  \textsc{3-LS}   & 0.66 & \textbf{0.91}  \\
  \textsc{4-LS}   & 0.62 & 0.83  \\
  \textsc{TMCD} & 0.36 & 0.60 \\
 \bottomrule
\end{tabular}
\caption{Pearson correlation between the easiness score of a split and the average model EM, shown for \textsc{$n$-LS} and TMCD, for COVR and Overnight (synthetic).}
\label{tab:mcd-comparison}
\end{table}

As discussed (\S\ref{subsec:alternative}), Maximum Compound Divergence (MCD) and its variation TMCD, are recently-proposed metrics for estimating the difficulty of a test set, while we measure difficulty at the instance level. To measure how well can local structures predict performance at the split level, we average the EM scores that the 4 models get on each split and use this average as the ``gold'' easiness of that split. We then average the easiness predictions of our decision rule across all instances in a split to obtain an easiness prediction for that split. For TMCD, we compute the compound divergence of each split (high compound divergence indicates a more difficult split, or lower easiness) following \newcite{shaw-etal-2021-compositional}, see details in App.\ref{app:mcd}.

We evaluate by measuring Pearson correlation between the predicted scores and gold scores. For TMCD we take the negative of the predicted score, such that for both methods a higher correlation is better. We show results for COVR and Overnight only, since the number of splits in these two datasets is large enough (124 and 50 splits respectively). The results, shown in Table~\ref{tab:mcd-comparison}, demonstrate that \textsc{$n$-LS} correlates better with the EM of models.% for any order $n$ of structures.

\subsection{Model Architecture Effect}

%We have shown that unobserved structures are a key factor in explaining what makes compositional generalization hard.
We now check if our proposed decision rule generalizes beyond Transformer-based seq2seq models.
To that end, we repeat our experiments with an LSTM \cite{HochSchm97} decoder with a copying mechanism \cite{gu-etal-2016-incorporating}, and BERT-Base as the encoder.\footnote{Accuracy with an LSTM encoder was too low even in an i.i.d split.}

Results are given in Table~\ref{tab:decison-rule-results-lstm}, showing that AUC scores with LSTM are close to the scores with a Transfomer.
% (except S2Q, see below)
This indicates that LSTMs mostly err due to unobserved local structures, similar to transformers. The only exception is S2Q that gets  much lower EM compared to Transformers and a large drop in AUC as well.
In addition, our decision rule works well even though the EM of the LSTM is lower (70.0 for COVR, 50.8/20.5 for Overnight synthetic and paraphrased, 54.1 for S2Q, and 64.4 for ATIS), indicating that the cases where the LSTM is wrong and the Transfomer is correct, are often when the easiness prediction is lower.

\begin{table}[!t]
\centering
\scriptsize
\begin{tabular}{lllll}
\toprule
{\bf Decision Rule}       & {\bf COVR}    & {\bf Overnight}  & {\bf S2Q}    & {\bf ATIS} \\
&  & syn / nat. & & \\
 \midrule
 
  \textsc{2-LS} (Trans.)& 93.3 & 84.6 / 67.9         & 78.9          & 75.8  \\
  \textsc{3-LS} (Trans.)& 93.1 & 91.0 / 74.8 & 81.6          & 78.7  \\
  \textsc{4-LS} (Trans.)& 92.1 & 88.0 / 78.4 & 79.9          & 81.4 \\
 \midrule
  
  \textsc{2-LS} (LSTM) & 80.0 & 77.3 / 66.5 & 56.8 & 71.8  \\
  \textsc{3-LS} (LSTM)& 81.6 & 84.2 / 71.0 & 60.7 & 76.9 \\
  \textsc{4-LS} (LSTM)& \textbf{82.6} & \textbf{85.8} / \textbf{72.8} & \textbf{60.9} & \textbf{79.6} \\
  \bottomrule
\end{tabular}
\caption{AUC scores of our decision rule across each dataset, for experiments with an LSTM-based decoder (bottom) compared to a Transfomer (top).}
\label{tab:decison-rule-results-lstm}
\end{table}

\section{Leveraging Local Structures}

We show we can take advantage of the insights presented to (1) create challenging compositional splits (\S\ref{subsec:n-ls-splits}) and (2) to improve data sampling efficiency for compositional generalization (\S\ref{subsec:efficient-sampling}).

\subsection{$n$-LS Splits}
\label{subsec:n-ls-splits}

We showed that unobserved local structures explain compositional generalization failures. Next, we test our conjecture from the opposite direction: we evaluate accuracy when testing on adversarial splits designed to contain unobserved local structures. 

Our goal is to create splits such that the similarity between any structure in the entire set of training programs and any structure in the set of test programs will be minimal. We do this by going over the set of all $n$-LSs, $\mathcal{S}$, in the set of all programs $\mathcal{P}$. For each $s \in \mathcal{S}$ we define the set $\mathcal{S}_u$ which contains all structures that are similar enough to $s$. We then attempt to create a new split such that its test set will contain all examples that have any of the structures in $\mathcal{S}_u$. See App.~\ref{subapp:n-ls-splits} for full details.

We test two variations: one where $\mathcal{S}$ includes all $2$-LS structures, and one where we ignore sibling relations (\textsc{2-LS-NoSib}). In addition, we create another set of splits where instead of setting $\mathcal{S}_u$ to include \emph{all} structures that are similar to $s$, we instead randomly sample \emph{half} of them, meaning some similar structures will exist in the training set (\textsc{2-LS-NoSib-Half}). 
For each configuration, we generate multiple splits (see App.~\ref{subapp:n-ls-splits}), and compute average accuracy across splits and models. For the template split, we use 5 random splits.

%We compare accuracy on up to 15 of our generated splits (see App.~\ref{subapp:n-ls-splits}) with five random  template splits.

Table~\ref{tab:ngram-split-results} shows that EM of models on our adversarial splits are dramatically lower than on template splits, especially for the \textsc{2-LS-NoSib} variation, with a difference of 74.9 or 45.8 absolute points. Moreover, the scores on \textsc{2-LS-NoSib-Half} are much higher, re-enforcing our hypothesis that unobserved structures are only hard if there are no similar observed structures.

\begin{table}[!t]
\centering
\scriptsize
\begin{tabular}{lll}
\toprule
{\bf Split}       & {\bf COVR} & {\bf Overnight} \\
 \midrule
  \textsc{2-LS}   & 32.8 $\pm$ 43.6    & 23.1 $\pm$ 33.6  \\
  \textsc{2-LS-NoSib} & \textbf{19.7} $\pm$ 34.6 & \textbf{8.4} $\pm$ 15.8 \\
  \textsc{Template} & 94.6 $\pm$ 21.7 & 54.2 $\pm$ 25.7\\
 \midrule
  \textsc{2-LS-NoSib-Half}   & 92.6 $\pm$ 20.0 &  38.6 $\pm$ 32.4 \\
 \bottomrule
\end{tabular}
\caption{EM on adversarial splits versus template splits. Numbers are averaged across all created splits and across 4 models. We show standard deviation across the EM scores of all models on all generated splits.}
\label{tab:ngram-split-results}
\end{table}

\subsection{Efficient Sampling}
\label{subsec:efficient-sampling}

We now test if we can leverage unobserved local structures to choose examples that will lead to better compositional generalization. We assume access to a large set of examples $\mathcal{D}_{\textrm{pool}}$, but we can only use a small subset $\mathcal{D}_{\textrm{train}} \subset \mathcal{D}_{\textrm{pool}}$ for fine-tuning, where the budget for training is $|\mathcal{D}_{\textrm{train}}| \leq B$. Our goal is to improve accuracy on an unseen compositional test set $\mathcal{D}_{\textrm{test}}$, by choosing examples that are likely to reduce the number of unobserved structures in the test set. To simulate this, we use the template split method to generate $\mathcal{D}_{\textrm{pool}}$ and $\mathcal{D}_{\textrm{test}}$ for COVR, ATIS and S2Q, holding out 20\% of the program templates for COVR and ATIS, and 70\% for S2Q (the number of templates in Overnight was too low to use). Improving compositional generalization under budget constraints was explored by \newcite{oren-etal-2021-finding}, but as our setup is different, results are not directly comparable.

We propose a simple iterative algorithm, starting with $\mathcal{D}_{\textrm{train}} = \phi$. We want our model to observe a variety of different local structures to increase the chance of seeing all local structures in $\mathcal{D}_{\textrm{test}}$. Thus, at each step, we add an example $e \in \mathcal{D}_{\textrm{pool}}$ that contains a local structure that is unobserved in $\mathcal{D}_{\textrm{train}}$. We do this by first sampling an unseen local structure (if all structures are already observed, we uniformly sample one), and then randomly picking an example that contains this structure. We continue until $|\mathcal{D}_{\textrm{train}}| = B$.

\begin{figure}
  \centering
  \includegraphics[width=0.8\linewidth]{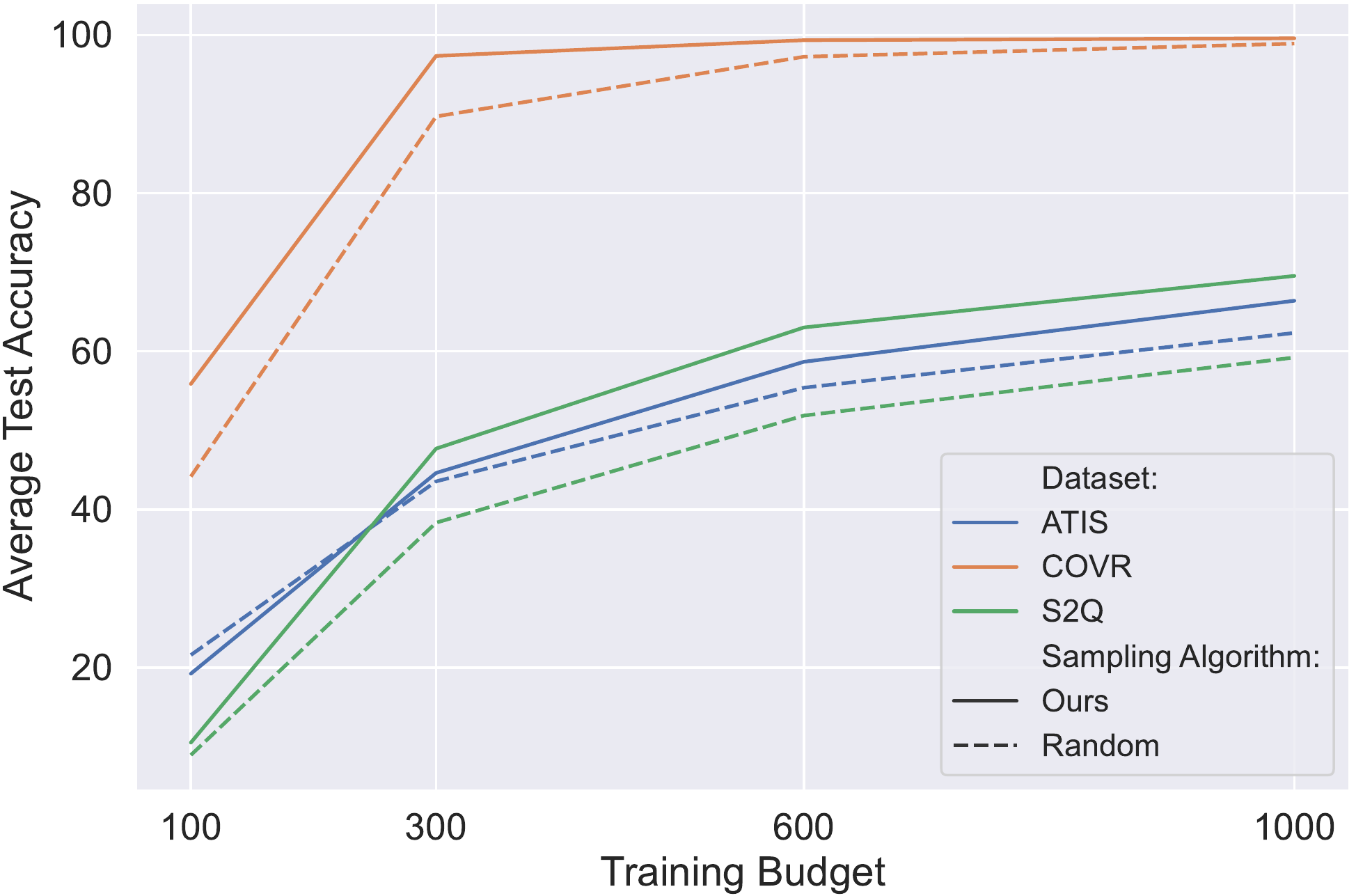}
  \caption{Average test accuracy when sampling based on local structures vs. random sampling on COVR, ATIS and S2Q as training budget is varied.} 
  \label{fig:efficient_all}
\end{figure}

We compare our algorithm against random sampling without replacement across a spectrum of training budgets. We evaluate both methods on 5 different template splits for each dataset, and for each template split, budget and sampling algorithm we perform the experiment with 3 different random seeds. Figure \ref{fig:efficient_all} shows the average compositional test accuracies on $\mathcal{D}_{\textrm{test}}$ for the two algorithms on the three datasets. Our sampling scheme outperforms or matches the random sampling method on almost all datasets and training budgets.
\section{Related Work}

\paragraph{Benchmarks}
Compositional splits are have been defined either manually \cite{lake2018scan,bastings-etal-2018-jump,Bahdanau2019CLOSUREAS,kim-linzen-2020-cogs,gscan2020,bogin-etal-2021-covr}  or automatically in a dataset-agnostic manner (as in \S~\ref{subsec:n-ls-splits}). Automatic methods include splitting by output length \cite{lake2018scan}, by anonymizing programs  \cite{finegan-dollak-etal-2018-improving}, and by maximizing divergence between the training and test sets \cite{keysers2020measuring,shaw-etal-2021-compositional}.

\tightparagraph{Improving generalization} Many approaches have been proposed to examine and improve generalization, including the effect of data size and architecture \cite{furrer2020Compositional}, data augmentation \cite{andreas-2020-good,akyrek2021learning,guo2021revisiting}, data sampling \cite{oren-etal-2021-finding}, model architecture \cite{herzig-berant-2021-span,bogin-etal-2021-latent,chen2020compositional}, intermediate representations \cite{herzig2021unlocking} and different training techniques \cite{oren-etal-2020-improving,csordas-etal-2021-devil}.

\tightparagraph{Measuring compositional difficulty} The most closely related methods to our work are
%which measures the predicted difficulty of a compositional split 
MCD and its variation TMCD \cite{keysers2020measuring,shaw-etal-2021-compositional}, designed to create compositional splits. Both methods define a tree over the program of a given instance. In MCD, it is created from the tree of derivation rules that generates the program, and in TMCD from the program parse tree, similar to our approach. Then, certain sub-trees are considered as compounds, which are analogous to $n$-LSs (with the main exception that $n$-LSs contain consecutive siblings). Splits are created to maximize the divergence between the distributions of compounds in the training and test sets. 

However, since MCD and TMCD were designed to generate compositional splits, they were not tested on whether they \emph{predict} difficulty of other splits, such as the template split. In addition, while in TMCD the difficulty is over an entire test set, we predict the difficulty of specific instances. Instance-level analysis can better characterize the challenges of compositional generalization, and as we show in \S\ref{subsec:mcd} it is a better predictor of difficulty compared to TMCD even at the split level. Moreover, our approach can also be used to create challenging compositional splits (\S\ref{subsec:n-ls-splits}). Another important distinction is that we introduce the concept of structure similarity, which further improves the ability to predict the difficulty of instances.
\pagebreak
\section{Discussion}
\label{sec:limitations}

\tightparagraph{Limitations}
In this work we only use the \emph{programs} of test instances to predict difficulty, but ignore the utterance. Thus, we do not take into account language variability, which could play a factor in compositional splits. 
In addition, we analyze datasets where models get high i.i.d accuracies. Our decision rule may not work when other difficulties conflate with the compositional challenge.

\tightparagraph{Conclusion}
We have shown that unobserved local structures have a critical role in explaining the difficulty of compositional generalization over a variety of datasets, formal languages, and model architectures, and demonstrated how these insights can be used for the evaluation of compositional generalization, and to improve data efficiency given a limited training budget. We hope our insights would be useful in future work for improving generalization in sequence to sequence models.

\section*{Acknowledgements}
This research was partially supported by The Yandex Initiative for Machine Learning and the European Research Council (ERC) under the European Union Horizons 2020 research and innovation programme (grant ERC DELPHI 802800). DARPA MCS program under Contract No. N660011924033 with the United States Office of Naval Research. NSF grant \#IIS-1817183. This work was completed in partial fulfillment for the Ph.D degree of Ben Bogin.

We thank Elad Segal for his code to measure compound divergence (MCD), and Jonathan Herzig, Ankit Gupta and Sam Bowman for their helpful feedback.

\bibliography{anthology,custom}
\bibliographystyle{acl_natbib}

\clearpage

\appendix
\section{Datasets}
\label{app:datasets}

In this appendix we enumerate the different preprocessing steps, anonymization functions and specific evaluation methods used for each dataset, if any, and provide the number of splits/instances created in each splitting method. Note that the described anonymization functions are used only in the template method splits, and are not used in the grammar splits or adversarial splits.

\subsection{COVR}

\paragraph{Generation}
The COVR dataset that we use is generated with a manually written grammar that is based on the visual question answering (VQA) dataset of the same name \cite{bogin-etal-2021-covr}, with utterances that contain either true/false statement or questions regarding different objects in an image. However, the dataset we use is separate from the VQA dataset since it generates only the utterances and the executable programs, without any dependence on a scene graph or image. While we use a grammar to generate the pairs, during generation some pre-defined pruning rules prevent specific cases where illogical programs are produced. The entire grammar, pruning rules and generation code are available in our codebase.

\paragraph{Anonymization}
We anonymize the following groups of symbols (symbols in each group are replaced with a group-specific constant): numbers, entities (\texttt{dog},\texttt{cat}, \texttt{mouse}, \texttt{animal}), relations (\texttt{chasing}, \texttt{playing\_with}, \texttt{looking\_at}), types of attributes (\texttt{color}, \texttt{shape}), attribute values (\texttt{black}, \texttt{white}, \texttt{brown}, \texttt{gray}, \texttt{round}, \texttt{square}, \texttt{triangle}) and logical operators (\texttt{and}, \texttt{or}).

\subsection{Overnight}
\paragraph{Preprocessing} We remove redundant parts of the programs, namely, the scope prefix that is used in functions and entities (\texttt{edu.stanford.nlp.sempre.overnight} \texttt{.SimpleWorld.XXX}) and declarations of types (\texttt{string}, \texttt{number}, \texttt{date}, \texttt{call}). The \emph{regex} dataset is not used due to parsing issues.

\paragraph{Anonymization}
We anonymize the following groups of symbols: strings, entities and numbers. We use the type declaration that are removed in the preprocessing part to identify these groups.

\paragraph{Evaluation}
When evaluating the synthetic versions of Overnight, we did not encounter any evaluation issues. However, on the paraphrased version, some of the instances yielded false negatives, mostly due to inconsistent order of filter conditions. For example, the program for the paraphrased utterance \emph{``find an 800 sq ft housing unit posted on january 2''} first defines the posting date as a condition, and only then the square feet size. This could happen whenever the crowdsource workers changed the order of the conditions as part of paraphrasing. In such cases, most of the times, models output conditions in the exact order given in the utterance, which would result in a negative exact match accuracy, even though the program is essentially correct. We normalize the order of the conditions to prevent such cases. We have encountered other evaluation issues as well which we did not address, since they were not as common.

\subsection{S2Q}
\paragraph{Preprocessing}
The raw generated S2Q dataset provided by \newcite{oren-etal-2021-finding} has several issues which made i.i.d results of different models to be too low (due to issues that are not related to compositionality, described below), and $n$-LSs to be sometimes non-meaningful (since the Thingtalk language is not entirely functional). Thus, we perform several preprocessing steps that were not necessary for the other dataset.

\begin{table*}[!t]
\centering
\footnotesize
\begin{tabular}{p{2.2cm}p{12.95cm}}
\toprule
  \makecell[l]{Orig. program} & \makecell[l]{\texttt{now => ( @org.schema.Person.Person ) filter count ( param:award:Array}\\\texttt{(String) ) >= 8 => notify}}\\
  
  Preprocessed & \makecell[l]{\texttt{filter ( Person ) ( count ( award ) >= NUMBER\_VAL )}}\\
  \midrule
  
  \makecell[l]{Orig. utterance} & \textit{which are the person which have either \textbf{learning of true university} or \textbf{ky.} in the works for and having job title containing \textbf{animal health technician} that have \textbf{alger lake , mi}}\\
  
  Anonymized & \textit{which are the person which have either \textbf{worksFor} or \textbf{worksFor} in the works for and having job title containing \textbf{jobTitle} that have \textbf{workLocation}}\\
 \bottomrule
\end{tabular}
\caption{An example for the preprocessing and anonymization we perform for the S2Q dataset. In the top example, we remove redundant parts, anonymize numbers and do minor modifications such that \texttt{filter} will be the function that calls the other arguments. In the bottom example, we anonymize the utterance to prevent ambiguity for the column type of the entity \textit{alger lage , mi} (see description in text). Entities are bold.
}
\label{tab:s2q-preprocessing}
\end{table*}

First, similar to Overnight, we remove redundant parts that define scope (e.g. \texttt{@org.schema.Person.Person} is replaced with \texttt{Person}). Next, we perform slight changes to the programs of S2Q such that their parsed tree better describes the hierarchy of the function and arguments. For example, see Tab.~\ref{tab:s2q-preprocessing} (top), where we replace the positions of \texttt{filter} and \texttt{Person}, such that \texttt{filter} will be the function that calls \texttt{Person} as its argument. Next, since S2Q programs were generated with thousands of different random entities names which are often interleaved in a non-natural way in the utterance (e.g. \textit{``people which are alumni of seems like some people , , with job title containing clarifier , or''}, where \textit{``seems like some people''} and \textit{``clarifier , or''} are entities), we anonymize these strings by replacing their occurrences in the program with a constant value \texttt{STR\_VAL} (we do so for numbers as well).
To anonymize the utterance, we do not replace string values with just a constant value, due to another related issue: in some cases, filter conditions are used in the program without the utterance mentioning the relevant column that should be filtered. For example, the phrase \textit{``which are the person ... that have alger lake , mi''} refers to people that have \textit{``alger lake , mi''} in the column \texttt{workLocation}, however, the usage of this specific column cannot be inferred from the name of the entity. We thus replace every string value in the utterance with the name of the column that it should use. See Tab.~\ref{tab:s2q-preprocessing} (bottom) for an example.
Note that unlike the other datasets, here anonymization is used not only for the cause of template splitting, but rather models are trained and evaluated with these anonymized utterances and programs. 

\paragraph{Anonymization}
String values and numbers are already anonymized, as described above. For the purpose of template splitting, we additionally replace the names of fields (e.g. \texttt{worksFor}, \texttt{alumniOf}, \texttt{faxNumber}, etc.) in the programs with a constant value, and the different operators (\texttt{==}, \texttt{>=}, \texttt{<=}, \texttt{~=}, \texttt{contains}, \texttt{asc} and \texttt{desc}) as well.

\paragraph{Evaluation}
We normalization S2Q programs during evaluation to prevent false negative cases, where the predicted program is different than the gold, but the two have the same meaning (i.e. would provide the same answer given any set of input). We address two specific cases. First, we normalize \texttt{and} and \texttt{or} clauses, such that
clauses are sorted by alphabetic order while taking into account the precedence between the two operators. Second, we remove a redundant call of the \texttt{compute count} function that is used twice and can thus be ommitted (see our codebase for exact implementation).

\subsection{ATIS}
\paragraph{Preprocessing} The ATIS dataset contains numbered variables (\texttt{\$0}, \texttt{\$1}, etc). However, it has some inconsistencies in the way these variables are given: sometimes the number appears directly after the dollar sign (\texttt{\$0}), and sometimes the letter ``v'' appears between them (\texttt{\$v0}). Additionally, we standardize the order of these numbers by setting the number $n_i$ of the $i$-th variable (where the order of variables is defined by the position of their first appearance in the program), will be equal to $n_{i-1} + 1$, except for the first variable, which is 0, so that variables will be numbered in a consistent manner.
\section{Splits}
\label{app:splits}

\subsection{Template Split}

The random template split, proposed in \citet{finegan-dollak-etal-2018-improving}, first converts programs into \emph{abstract templates} using a program anonymization function, that replaces certain program tokens such as string values, numbers and entities with their abstract type (e.g., for the input \textit{``are there people that works for crossbars~?''} in Table~\ref{tab:datasets}, the string value \texttt{crossbars} is replaced with \texttt{STR\_VAR}). See App.~\ref{app:datasets} for the exact anonymization function used in each dataset.

We group examples according to their abstract template, randomly split the templates into a training set and a test set, and place each example in the train/test set according to their abstract template. While not a part of the original procedure, to make sure splits are ``solvable'', we verify in each split that there is no token that appears in the test set but does not appear in the training set (otherwise, we discard the split). To obtain multiple splits, we perform multiple random splits of templates.

To make sure the distribution of number of examples per template is not very skewed, we limit the number of examples for each program template to $k_{\textrm{train}}=1000$ in the training set.
In case there are more, we randomly sample $k_{\textrm{train}}$ examples for that template and discard the rest. We follow the same procedure for the test split, with $k_{\textrm{test}}=10$. 

\subsection{Grammar Split}
\label{subapp:grammar}
We propose a grammar split, which can be used when we have the set of context-free grammar rules $\mathcal{G}$ that generate the dataset programs. This can create meaningful splits, since generation is not random. For example, this can create a split where the test set contains exactly the set of examples that have the token \texttt{and} as a parent of the token \texttt{exists} (in the program tree). Such a split is unlikely to emerge in a template split.
%(see App.~\ref{subapp:grammar} for example grammar splits).

To create a grammar split for a set of examples, we look at the \emph{derivation} $d = (r_1, ..., r_{N_d})$ of each example program, i.e., the sequence of production rules from the grammar that generate that program, where $r_i \in \mathcal{G}$. We create a compositional split by holding out a small number $N_U$ of \emph{pairs} of grammar rules, $\mathcal{U}=\{(r_{1_i}, r_{2_i})\}_{i=1}^{N_\mathcal{U}}$, and define the test set to contain any example whose program derivation contains a pair of rules $r_1$ and $r_2$, where $(r_1, r_2) \in \mathcal{U}$.
For example, if $\mathcal{U}$ contains a single pair of rules $(r_1, r_2)$, where $r_1$ is a rule that produces the terminal \texttt{and} and $r_2$ is a rule that produces the terminal \texttt{exists}, then the test set will include only (and all) examples that have both these rules in their derivation. Due to the structure of the grammar, this means the test programs will always have \texttt{exists} as a child of \texttt{and} in the program tree. Importantly, the training set will still have examples with both of these two terminals separately.
We repeat this process with different instances of $\mathcal{U}$, as we desribe next.

\begin{figure*}
  \centering
  \includegraphics[width=\linewidth]{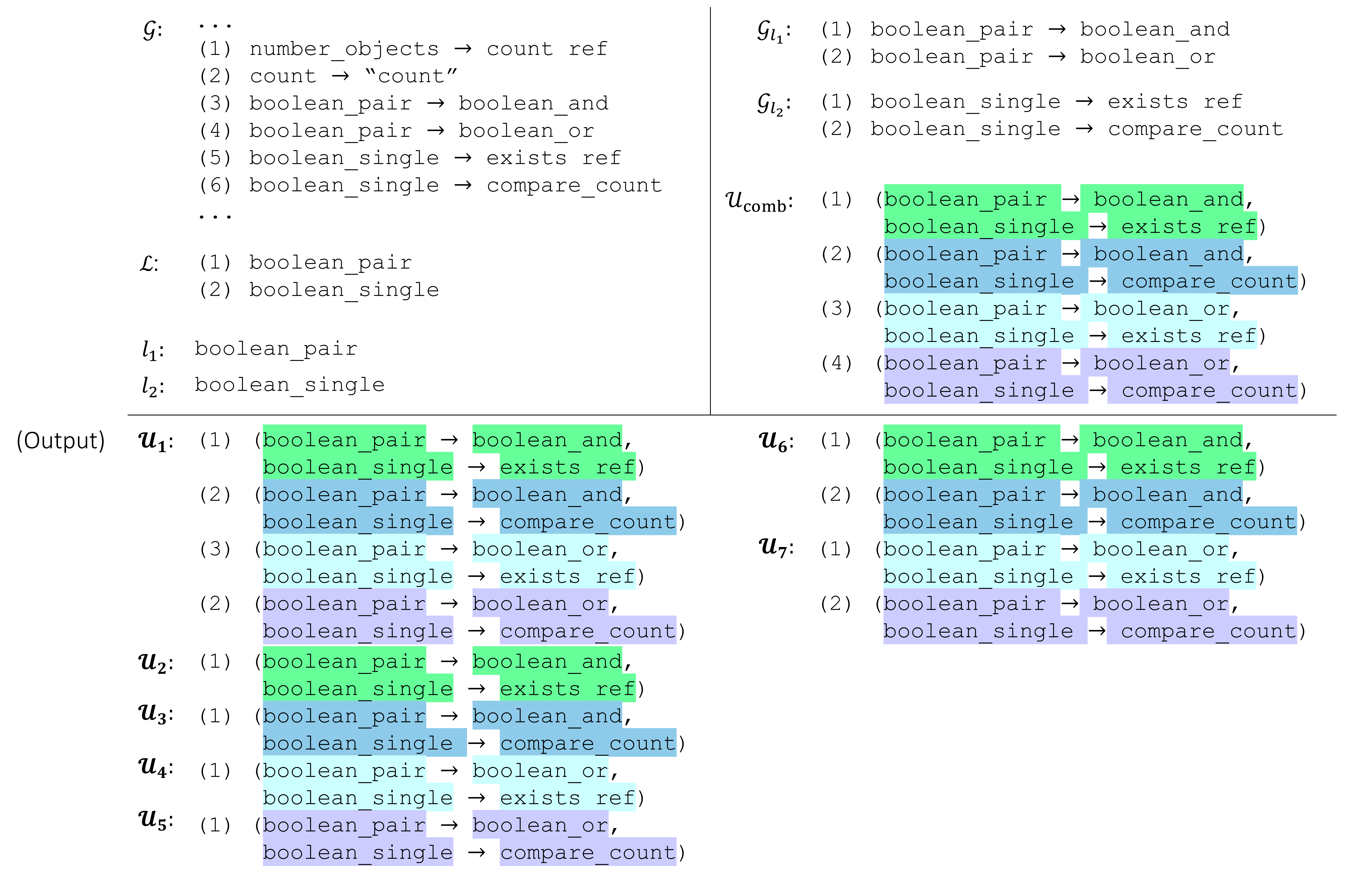}
  \caption{An illustration of our grammar split. Bottom part shows 7 produced different splits. See text for details.} 
  \label{fig:grammar-split}
\end{figure*}

A grammar rule $r_i=A \rightarrow \gamma \in \mathcal{G}$ is comprised of its left hand side (LHS) $A$, which is a non-terminal, and its right hand side (RHS) $\gamma$ which is a sequence of terminals and/or non-terminals. In theory, we could have iterated over all possible \emph{sets} of pairs, and create a split out of each of these sets, but this would be impractical. Instead, we propose the following method to to pick different instances of $\mathcal{U}$.

First, we consider only the set $\mathcal{L}$ of non-terminals that are \emph{``meaningful''}. We consider a non-terminal $l$ to be meaningful iff (1) there are at least two rules in $\mathcal{G}$ where $l$ is their LHS, or (2) $l$ is a non-terminal such that for at least two different rules $r_a, r_b \in \mathcal{G}$, $l$ belongs to the RHS sequence of both $r_a$ and $r_b$. By selecting only meaningful non-terminals, we can avoid considering redundant rules that always appear together with other rules. For example, in Fig.~\ref{fig:grammar-split}, the rule \texttt{count} is a non-terminal of just a single rule, and it belongs to the RHS of just a single rule, thus it is not considered meaningful. In this example, $\mathcal{L}$ will consist of the meaningful non-terminals \texttt{boolean\_pair} and \texttt{boolean\_single}.

We then iterate over all possible unique pairs $(l_1, l_2)$ of non-terminals in $\mathcal{L}$.  For each such pair, we take the set $\mathcal{G}_{l_1}$ of grammar rules for which $l_1$ is the LHS of, and the set $\mathcal{G}_{l_2}$ of grammar rules for which $l_2$ is the LHS of. In the example in Fig.~\ref{fig:grammar-split}, we will have one possible such pair of non-terminals $(l_1, l_2)$, \texttt{boolean\_pair} and \texttt{boolean\_single}. The figure shows the set $\mathcal{G}_{l_1}$ that contains all rules with \texttt{boolean\_pair} as its non-terminal, and similarly for $\mathcal{G}_{l_2}$.

Given the two sets of rules $\mathcal{G}_{l_1}$ and $\mathcal{G}_{l_2}$, we set $\mathcal{U}_\textrm{comb}$ to be the set of all possible combinations from $\mathcal{G}_{l_1}$ and from $\mathcal{G}_{l_2}$ (the Cartesian product of the two sets). Next, we use the following pseudo code to generate different instances of $\mathcal{U}$ (each set that we yield is a different generated instance of $\mathcal{U}$):
\begin{algorithm}
\begin{algorithmic}[1]
  \STATE yield $\mathcal{U}_\textrm{comb}$ 
  \FOR {$(r_1, r_2) \in \mathcal{U}_\textrm{comb}$}
  \STATE yield $\{(r_1, r_2)\}$
  \ENDFOR
  \FOR {$r \in \left( \mathcal{G}_{l_1} \cup \mathcal{G}_{l_2} \right)$}
  \STATE yield \{$(r_1, r_2) \in \mathcal{U}_\textrm{comb}$ if $r_1=r$ or $r_2=r$\}
  \ENDFOR
\end{algorithmic}
\end{algorithm}

Following the figure, $\mathcal{U}_6$ will be generated in line~1, $\mathcal{U}_\textrm{2..5}$ in line~3, and $\mathcal{U}_{6..7}$ in line~6. Finally, for each instance of generated $\mathcal{U}$, we discard it if it generates an invalid split (where symbols in the test set do not occur in the training set).

\begin{table*}[!t]
\centering
\footnotesize
\begin{tabular}{p{2.2cm}p{12.95cm}}
\toprule
  \makecell[l]{$\mathcal{U}$} & \makecell[l]{
  1. (\texttt{eq $\rightarrow$ 'eq'}, \texttt{boolean\_pair $\rightarrow$ boolean\_or})\\
  2. (\texttt{eq $\rightarrow$ 'eq'}, \texttt{boolean\_pair $\rightarrow$ boolean\_and})
  }\\
  \hdashline
  Train examples & \makecell[l]{
1. \textit{either there is black cat or all of mouse are looking at animal} \\
\texttt{\textbf{or}(exists(filter(black,find(cat))),all(find(mouse),with\_relation}\\\texttt{(scene(),looking\_at,find(animal))))} \\
2. \textit{the number of black dog is equal to the number of white cat} \\
\texttt{\textbf{eq}(count(filter(black,find(dog))),count(filter(white,find(cat))))}
}\\
  \hdashline
  Test example & \makecell[l]{
\textit{either the color of cat that is playing with cat is equal to black or there is mouse} \\
\texttt{\textbf{or}(\textbf{eq}(query\_attr[color](with\_relation(find(cat),playing\_with,find(cat))),}\\\texttt{black),exists(find(mouse)))}
  }\\
  \hdashline
  \makecell[l]{\\Predicted easiness\\for split (\textsc{2-LS})} & \makecell[l]{0.09}\\
  \midrule
  
  %%%%%%%%%%%%%%%%%%%%%%%%%%%%

  \makecell[l]{$\mathcal{U}$} & \makecell[l]{
  1. (\texttt{ref $\rightarrow$ with\_relation}, \texttt{ref $\rightarrow$ filter\_object})
  }\\
  \hdashline
  Train examples & \makecell[l]{
1. \textit{the number of white cat is less than the number of gray mouse} \\
\texttt{lt(count(\textbf{filter}(white,find(cat))),count(\textbf{filter}(gray,find(mouse))))} \\
2. \textit{is the color of cat that is looking at animal black or white ?} \\
\texttt{choose(query\_attr[color](\textbf{with\_relation}(find(cat),looking\_at}\\\texttt{,find(animal))),black,white)}
}\\
  \hdashline
  Test example & \makecell[l]{
\textit{the number of animal that is looking at white animal is greater than 4} \\
\texttt{gt(count(\textbf{with\_relation}(find(animal),looking\_at,\textbf{filter}(white,}\\\texttt{find(animal)))),4)}
  }\\
  \hdashline
  \makecell[l]{\\Predicted easiness\\for split (\textsc{2-LS})} & \makecell[l]{0.43}\\
  \midrule
  
  %%%%%%%%%%%%%%%%%%%%%%%%%%%%

  \makecell[l]{$\mathcal{U}$} & \makecell[l]{
  1. (\texttt{boolean\_single $\rightarrow$ exists ref}, \texttt{boolean\_pair $\rightarrow$ boolean\_or})\\
  }\\
  \hdashline
  Train examples & \makecell[l]{
1. \textit{either some of animal are brown or the number of cat is less than 2} \\
\texttt{\textbf{or}(some(find(animal),filter(brown,scene())),lt(count(find(cat)),2))} \\
2. \textit{both the number of dog that is chasing cat that is chasing dog is equal to 4 and there is brown mouse}\\\textit{that is chasing mouse} \\
\texttt{and(eq(...),\textbf{exists}(with\_relation(filter(brown,find(mouse)),chasing,}\\\texttt{find(mouse))))
}
}\\
  \hdashline
  Test example & \makecell[l]{
\textit{either most of round white mouse are square or there is animal that is playing with cat} \\
\texttt{\textbf{or}(most(...),\textbf{exists}(with\_relation(find(animal),playing\_with}\\\texttt{,find(cat))))}
  }\\
  \hdashline
  \makecell[l]{\\Predicted easiness\\for split (\textsc{2-LS})} & \makecell[l]{0.94}\\

 \bottomrule
\end{tabular}
\caption{Selected splits generated with the grammar split method, with selected examples for each split.}
\label{tab:example-grammar-splits}
\end{table*}

\paragraph{Example grammar splits} We show a few selected examples of the generated grammar splits in Table~\ref{tab:example-grammar-splits}.

\begin{table*}[!t]
\centering
\footnotesize
\begin{tabular}{llllll}
\toprule
{\bf Dataset} & {\bf Split} & {\bf \# Splits} & {\bf \# Total instances} & {\bf Avg \# training templates / ex.} & {\bf Avg \# test templates / ex.} \\
 \midrule
  COVR & Grammar & 124 & 62,000 & 1,510.0 / 3000.0 & 372.5 / 500.0  \\
  Overnight & Template & 5 & 2,707 & 43.9 / 1,491.8 & 10.8 / 215.0  \\
  S2Q & Template & 5 & 4,089 & 42.0 / 15,940.2 & 97.0 / 831.6   \\
  Atis & Template & 5 & 2,587 & 920.0 / 4,026.2 & 229.0 / 772.8  \\
 \bottomrule
\end{tabular}
\caption{Different splits statistics, including the number of generated splits per each split method and dataset, the total number of test instances across all splits and average train/test templates and examples across the splits.}
\label{tab:splits-stats}
\end{table*}

\subsection{Sizes of Splits}
\label{app-sub:split-sizes}
For Overnight and ATIS, we hold out 20\% of all program templates. For Schema2QA, we hold out 70\%, since otherwise model performance is too high.

For each dataset, we list the number of splits, the total number of instances over which we compute AUC scores, and the average number of train/test templates (after anonymization) in Table~\ref{tab:splits-stats}.

\begin{figure*}
  \centering
  \includegraphics[width=0.9\linewidth]{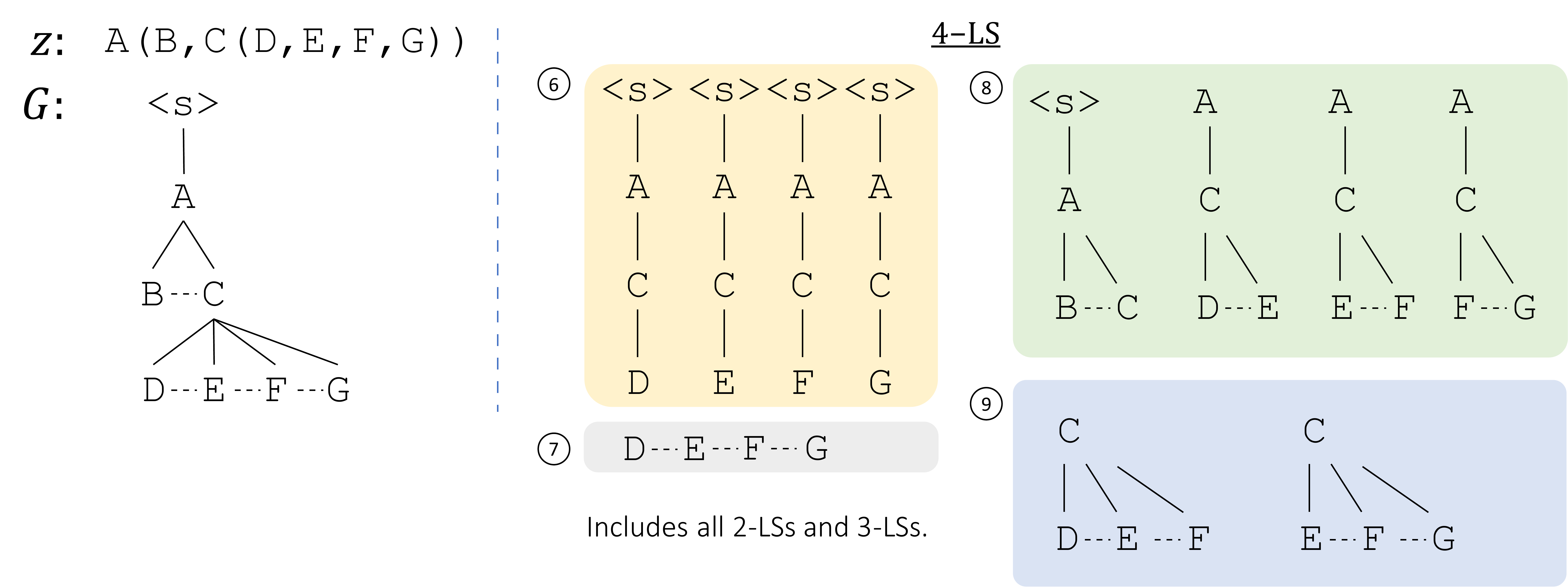}
  \caption{An example program $z$ and the structure of its program graph $G$ (left), with solid edges for parent-child relations and dashed edges for consecutive siblings. The right side enumerates all 4-LS structures over this graph.}
  \label{fig:4-ls}
\end{figure*}

\subsection{$n$-LS Splits}
\label{subapp:n-ls-splits}
As we mention in \S\ref{subsec:n-ls-splits}, we create adversarial splits by leveraging $n$-LSs, then evaluate the accuracy of models on these spits  against accuracy on a random template split, where we hold out $K=0.3$ of the program templates.

To create an adversarial split, we want to split the entire set of programs $\mathcal{P}$ into a set of training programs $\mathcal{P}_o$ and test programs $\mathcal{P}_u$, such that
if we look at the set $\mathcal{S}_o$ of observed $n$-LSs in $\mathcal{P}_o$, and the set $\mathcal{S}_u$ of unobserved $n$-LSs in $\mathcal{P}_u$, the similarity between any structure in $\mathcal{S}_o$ and any structure in $\mathcal{S}_u$ will be minimal. Given $\mathcal{S}_u$, a split of examples is created by setting the test set to contain all examples that have any of the structures in $\mathcal{S}_u$, and the training set to contain all other examples.

Concretely, to create an adversarial split, we go over the set of all $n$-LSs, $\mathcal{S}$, in the set of programs $\mathcal{P}$, and from each $s \in \mathcal{S}$ we attempt to create an adversarial split independently. Given an $n$-LS, $s$, and a similarity threshold $t$, a candidate set of unobserved structures is $\mathcal{S}_u(t) = \{ s_u \in \mathcal{S} \textrm{ s.t. sim}(s_u, s) > t \}$, that is, the set of all structures that are similar to $s$ (which includes $s$ by definition). 
When $t$ is low, splits are hard, since structures in $\mathcal{S}_o$ will be less similar to $\mathcal{S}_u$. On the other hand, when $t$ is too low, we might hold out a large fraction of programs, which might (a) create invalid splits, where symbols in the test set do not occur in the training set, or (b) the fraction of test set programs will exceed $K$, rendering comparison to a random template split unfair. Thus, we pick $\mathcal{S}_u$ with the lowest $t$ such that (1) $\mathcal{S}_u(t)$ is valid (all test symbols occur in the training set) and (2) the fraction of program templates in the test set is at most $K$. We merge identical splits that are created from different structures $s,s' \in \mathcal{S}$. In addition, because our process does not guarantee a difficult split, we discard any split if its easiness score is higher than a threshold $\tau$, as described next.

We choose $\tau$ by using the $n$-LS classifiers we evaluate in \S\ref{subsec:results}, for which we have shown AUC scores in Table~\ref{tab:decison-rule-results}. For each decision rule and for each dataset, we find an optimal threshold that optimizes the $F_1$ score on our test instances, and use it as the threshold $\tau$ for the matching $n$-LS split.

For \textsc{2-LS}, COVR, $\tau=0.58$ (11 splits generated, none discarded) and for Overnight, $\tau=0.13$ (112 splits are generated, 38 kept after filtering, 15 are sampled for experiments). For \textsc{2-LS-NoSib}, COVR, $\tau=0.3$ (5 splits generated, 1 split is discarded) and for Overnight, $\tau=0.2$ (44 splits are generated, 11 kept after filtering). For \textsc{2-LS-NoSib-Half}, COVR, we do not use $\tau$, but instead we take the top 15 hardest splits, according to the easiness score. We cannot use $\tau$ in these cases since the easiness score for all splits were higher than the thresholds (by the definition of our decision rule, easiness score is high in splits where all unobserved structures have similar observed structures).
\section{Training}
\label{app:training}
All models are fine-tuned on each of the generated splits, for a total of 64 epochs with a batch-size of 28 (except for experiments with T5-large, for which we've used a batch-size of 8). We use a learning rate of $3e^{-5}$ with polynomial decay. Each experiment was run with an Nvidia Titan RTX GPU.

We do early stopping using the test set, since we do not have a development set. As our goal is not to improve performance but only analyze instance difficulty, we argue this is an acceptable choice in our setting.
\section{4-LS}
\label{app:4-ls}

In \S\ref{subsec:structures}, we define 2-LS and 3-LS over the program graph $G$. The structure 4-LS is a natural extension of these structures. It includes all 2-LSs, and 3-LS, and also structures with (1) three parent-child relations, (2) three siblings relations and (3) a grandparent with its child and two sibling grandchildren and (4) a parent with three siblings (structures 6, 7, 8 and 9 in Figure~\ref{fig:4-ls}, respectively).
\section{Measuring TMCD}
\label{app:mcd}

We measure compound divergence of the distributions of compounds and atoms on the program graph, following \newcite{keysers2020measuring} and \newcite{shaw-etal-2021-compositional}. Similarly to TMCD, we define atoms and compounds over the program tree $T$, over the graph defined in \S\ref{subsec:structures}. Each tree node is considered an atom, and the compounds are all sub-trees of up to depth 2. We use the same Chernoff coefficient as in the original paper, $\alpha=0.5$, to compute compound divergence.

\end{document}